# A Novel Enhanced Convolution Neural Network with Extreme Learning Machine: Facial Emotional Recognition in Psychology Practices

Nitesh Banskota[1], Abeer Alsadoon[1,2,3,4*], P.W.C. Prasad[1,2,3], Ahmed Dawoud[1,2], Tarik A. Rashid[5], Omar Hisham Alsadoon[6]

[1] School of Computing Mathematics and Engineering, Charles Sturt University (CSU), Australia
[2] School of Computer Data and Mathematical Sciences, Western Sydney University (WSU), Sydney, Australia
[3] Kent Institute Australia, Sydney, Australia
[4] Asia Pacific International College (APIC), Sydney, Australia
[5] Computer Science and Engineering, University of Kurdistan Hewler, Erbil, KRG, IRAQ
[6] Department of Islamic Sciences, Al Iraqia University, Baghdad, Iraq

* Corresponding author. Dr. Abeer Alsadoon, Charles Sturt University, Sydney Campus, Sydney, Australia, Email: alsadoon.abeer@gmail.com , Phone +61 413971627

**ABSTRACT**

Facial emotional recognition is one of the essential tools used by recognition psychology to diagnose patients. Face and facial emotional recognition are areas where machine learning is excelling. Facial Emotion Recognition in an unconstrained environment is an open challenge for digital image processing due to different environments, such as lighting conditions, pose variation, yaw motion, and occlusions. Deep learning approaches have shown significant improvements in image recognition. However, accuracy and time still need improvements. This research aims to improve facial emotion recognition accuracy during the training session and reduce processing time using a modified Convolution Neural Network Enhanced with Extreme Learning Machine (CNNEELM). The proposed system consists of an optical flow estimation technique that detects the motion of change in facial expression and extracts peak images from video frames for image pre-processing. The system entails (CNNEELM) improving the accuracy in image registration during the training session. Furthermore, the system recognizes six facial emotions – happy, sad, disgust, fear, surprise, and neutral with the proposed CNNEELM model. The study shows that the overall facial emotion recognition accuracy is improved by 2% than the state of art solutions with a modified Stochastic Gradient Descent (SGD) technique. With the Extreme Learning Machine (ELM) classifier, the processing time is brought down to 65ms from 113ms, which can smoothly classify each frame from a video clip at 20fps. With the pre-trained InceptionV3 model, the proposed CNNEELM model is trained with JAFFE, CK+, and FER2013 expression datasets. The simulation results show significant improvements in accuracy and processing time, making the model suitable for the video analysis process. Besides, the study solves the issue of the large processing time required to process the facial images.

## 1. INTRODUCTION

Facial emotions convey a large amount of unspeakable information that could be an essential tool used by psychology prpractitionersuring clinical diagnosis [1, 2]. Human facial expressions are considered as a leading carrier to convey human emotion in communications. A study on non-verbal communication discloses that about 55% of human emotions are conveyed through facial expressions [1]. Human facial expression provides an important clue to understand and analyze human emotion and their behaviors. In recent times, several research in emotion analysis have been carried out extensively, and different methods have been deployed to recognize facial expressions for various applications such as driver monitoring, customer service, digital entertainment, and emotion robots [2-4].

In traditional Facial Emotion Recognition (FER) approaches, first, the face and its components are detected, then, the spatial and temporal features are extracted by handcrafted methods, and finally classified with a pre-trained facial expression classifier. Even though these methods of implementation occur instantaneously, the recognition accuracy is low due to the local variation (pose, illumination, alignment, and occlusion) of facial components that affects the facial expression feature's vectors. Commonly in the traditional approach, machine learning techniques such as Completed Local Binary Patterns (CLBP), Local Ternary Patterns (LTP), Local Phase Quantization (LPQ), Rotation Invariant Co-occurrence (RICo), and Rotated Local Binary Pattern Image (RLBPI) are used for feature's extraction. However, Support Vector Machine (SVM), AdaBoost, and random forest are used for classification.The traditional approaches require comparatively low memory and computational power; however, they require a large set of samples for training and various image pre-processing techniques. In recent years, traditional neural networking techniques are being replaced with a deep learning model, which has transformed the conventional way of machine learning. The multi-layer structure that forms a filter collection in the deep learning model differs from traditional machine learning [5]. This technological model provides an advantage over conventional image processing frameworks by creating a complex network for solving image pre-processing, feature extraction and classification [6]. Multiple deep network layers, commonly known as Convolution Neural Network (CNN) layers, act as a feature extractor that is independent of any classification task [7]. Thus, features are learned directly by observing the input image [8]. Though Convolution neural networking requires a large training dataset and high-performance computational power (GPUs and CPUs), its ability to automatically extracts features from images made it robust and effective.

Deep neural networks have been utilized extensively for facial emotion recognition—significant improvements achieved by using CNN and Deep belief networks DBN [10]. The recent studies carried on facial emotion recognition using various models of CNNs have achieved high accuracy of over 95% on emotion recognition using commonly available datasets like CK+, BU-3DEF and LFW datasets [10]. However, in these studies, the processing time required for computation has been ignored. The lowest processing time observed in the current state of the art solution is 135ms [10]. The contribution of this paper can be summarized as below,

- Proposes a model called Convolutional Neural network Enhanced with Extreme Learning Machine (CNNEELM) classifier.
- Our approach has proven improvements in the accuracy of ≈ 97% and reduced the processing time by half.
- The proposed model recognizes the facial expression from an image or a video clip in the unconstrained environment at high accuracy and low processing time for tele-psychological practices in real-time. Convolution Neural Network enhanced with Extreme Learning Machine (CNNEELM) model aims to extract salient facial patches with saliency detection algorithms. These patches minimized image aberration like illumination, low image resolution, occlusion, etc., with Extreme Learning Machine (ELM) classifier for faster image classification.

## 2. LITERATURE REVIEW

Facial Expression Recognition is a crucial human ability to determine how one person's expression conveyed the message of their emotion to another person. Research on developing such a computerized system similar to human perception are being carried out extensively [11]. However, in most of the studies, successful results are obtained under a constrained environment. In the last few years, numerous algorithms have been developed by researchers and have been suggested for FER. In these methodologies, the commonalities in most of the approaches are facial component detection, feature extraction, and expression classification. Several researchers have noticed that adopting the concept of Deep Learning with a convolution neural network model has better recognition and classification rate than the traditional machine learning techniques [2-4]. Many researchers have proposed their own models to improve the overall accuracy of state-of-art systems so that a highly reliable system can be introduced to recognize facial emotion from human expression. In the below sections, a literature review of various authors is observed based on several stages, which are: pre-processing, feature extraction, and classification.

### 2.1. Pre-processing

Wu et al. [12] examined how to customize the generic model when label information is not available in testing samples that causing a degradation in performance for emotion recognition system in terms of accuracy. In order to solve this problem, they used a feature mapping technique called Weighted Center Regression Adaptive Feature Mapping (W-CR-AFM) that adapts to the feature of new samples, which do not have labeled information, and correcting some misclassified samples; With their technique, they improve the recognition accuracy by about 3.01%, 0.49%, and 5.33% using Cohn-Kanade (CK+), Radboud Faces Database (RaFD), and Amsterdam Dynamic Facial Expression Set (ADFES) respectively. Additionally, they pre-processed testing samples with spatial normalization technique to straighten tilted head position that can contribute CNN model to achieve better training accuracy. Their mapping technique learns the data in batch by batch fashion rather than processing all training sets and validation set in a single batch. Hence, this technique does not offer further enhancement. Liu et al. [10] investigated the influence of various types of image distortion, such as illumination, occlusion, low image resolution, etc., in facial emotion recognition. In order to solve this problem, they used the Conditional Convolutional Neural Network Enhanced Random Forest (CoNERF) method to detect the salient feature in a facial image that eliminating the image distortions. The average accuracy of 94.09% on the multi-view BU-3DEF dataset, 99.02% on CK+ and JAFFE frontal facial datasets, and 60.9% on the LFW dataset were achieved. However, the prediction time of the method is more than 0.05 sec per frame for real-time video analysis, which is must be reduced for better efficiency. This method can be enhanced by using optical flow estimation to improve the quality of a saliency map so that only the required area of the facial image is made ready for system training. Li et al. [13] investigated the challenges that have to be faced while optimally combining the facial images of 2D and 3D face information for expression prediction to solve the challenging problems of illumination and pose variations. They proposed a solution to this problem by introducing learning-based feature-level fusion between SVM and Deep CNN. Therefore, the weights of 2D and 3D facial representations can be combined at the optimal size for multi-modal FER. Their solution is achieved the overall average recognition accuracy of 78.9% using two 3D face datasets which are BU-3DFE and Bosphorus. On the other hand, fusing the features between 2D and 3D faces can generate redundant features, which increases the processing time. This model can be enhanced by initializing the system with pre-trained CNN model and modifying the loss function for optimal training.

## 2.2. Feature Extraction

Zhang et al. [4] investigated that available algorithms for facial expression recognition is developed only for frontal-facial image and are not able to detect emotions from the facial expression for non-frontal FER. TO address this problem, they used a deep convolution neural network model that extracts facial features from multiple angle images varying from 0 to 45 degrees with 75% accuracy using a non-frontal facial expression database called BU-3DFE. Extraction of features with Scale-Invariant Feature Detection (SIFT) descriptors can be beneficial in their system since only the desired facial area can be represented with descriptors, and other irrelevant background image data can be discarded. However, only the frontal face has been considered, and no tests are performed for multi-view face images. Jain et al. [14] examined and compared the accuracy in facial emotion recognition in images with conventional sate-of-the-art-methods. They combined Convolution Neural Network CCN with Recurrent Neural Network RNN to extract the relation with facial images and temporal dependencies in images. The obtained class accuracy for CNN-RNN at 89.13% using the MMI dataset is 15% better than conventional CNN. This method for feature extraction does not require further refinement. While the accuracy of the frontal face image is high, no test has been conducted for the side face-view. The accuracy level may drop down for the side-angle facial images. Yao et al. [15] examined the challenge of coping with non-frontal head poses during facial expression recognition results in a considerable reduction of accuracy and robustness when capturing expressions that occur during natural communications. They offered a solution to this problem by introducing A depth-patch perceptron network (DPPNet) based 4D expression representation model to detect only peak frames with distinct facial emotion. The peak frames are fed to the pre-trained CNN model during the training session, which produced recognition accuracy of 64.7% on the BU4D-FE dataset. The recognition accuracy can be refined more with pre-processing of peak frames where only salient features of the face are detected and used to train the system. However, while processing only the peak frame, the frame with a micro-expression can be missed.

Peng et al. [16] investigated that emotions from facial expressions can be misleading as someone may try to deceive others by showing opposite facial expressions. In solution to this problem, they used dual temporal scale CNN for Micro-Expression Recognition, which analyses frames in a video clip for unusual emotion expression detected via a quick change in expression between frames. Average classification accuracy of 66.67% was achieved by this system using spontaneous micro-expression databases (CASME, I/II). However, the processing time of the system is increased, and feature points could be in the wrong position if the testing facial images have colour variation. Furthermore, the fusion of this method with any other method is only going to increase the processing time. Hence, additional techniques do not offer a further area of improvement. et al. [17] evaluated the need for accurate recognition of micro-expression for lie detection without being in physical contact with an individual. To address this situation, they researched on the micro-expression recognition system with small sample size frame by transferring the Long-term Convolutional Neural Network (TLCNN) which improved the recognition accuracy of state of art by 7% using 560 micro-expression video clips. TLCNN uses Deep CNN to extract features from each frame of micro-expression video clips, then feeds them to Long Short-Term Memory (LSTM) which learn the temporal sequence information of micro-expression. The reduction of mini-batch size during the training can refine the system with improved training accuracy. However, no further improvement in accuracy could be achieved. Ahmed et al. [18] investigated the impact of constraints like pose dissimilarity, age, lighting conditions and occlusions that minimizes the accuracy for facial emotion recognition system in a wild situation. They offer a solution to this problem by implementing an incremental active learning framework with CNN on wild facial expression with pre-trained VGG16 model. Besides, the system used new data labeling using the total active learning method to reduce the network training time. The system achieved an overall accuracy of 72.9% using the Intelligent Technology Lab (ITLab) dataset. However, the system fails to identify the micro-expression. Hence, the system can be enhanced by adding additional convolution layers for generalized feature extraction in VGG model.

### 2.3. Classification

Ruiz-Garcia et al. [19] examined that the socially assistive robots in the modern world that require high-level image processing with an accurate result at low processing time. In order to address this problem, they combined Deep CNN as a feature extractor with SVM as a classifier to classify the human in seven basic human emotions for improved human-robot interaction. With their system implementation, they could achieve 96.26% accurate results on the Karolinska Directed Emotional Faces (KDFE) dataset. However, the accuracy dropped down to 68.78% when used in the real world. This performance issue can be resolved by balancing the histogram of an input image and by using a better camera for capturing images. Kaya et al. [20] examined the large variance in wild Facial Emotion Recognition that occurred due to uneven illumination, cluttered background as well as a sudden change in facial expression for multimodal audio-visual recognition resulting in audio lag. In order to improve processing time, they used Extreme Learning Machine (ELM) fused with Partial Least Square regression (well known for fast learning capabilities) as a classifier. Their method has resulted in a decrease in processing time by 42.5%.

Moreover, the obtained accuracy of the proposed method is 98.47% on CK+ dataset and 72.46% on MMI dataset respectively. This method can be refined more by combining this work with the system proposed by Jain et al. [14] for highly accurate facial expression recognition. Liu et al. [10] investigated how actual word application's factors such as head pose variation, occlusion, and poor image quality can drop the recognition accuracy of the FER system and how the system requires very large training data to train the model. To address these problems, they fused Random Forest with Conditional CNN as a feature extractor and used their state-of-art NCFS function to improve the learning capability of the system. Conditional CoNERF is devised to enhance decision trees with the capacity of representation learning from transferred convolutional neural networks and to model facial expressions of different perspectives with conditional probabilistic learning. Their system achieved an accuracy of 98.8% on CK+ dataset and 84.48% on JAFFE datasets. The number of training samples is highly reduced with NCFS function without compromising the recognition accuracy. The processing time of this system is around 113ms. The enhancement of this system could be the utilization of Extreme Learning Machine (ELM) classifier to classify the extracted features which reduce the processing time due to less number of hidden layer in this classifier. Though they have achieved high accuracy for still images, this model does not process real-time videos.

### 3. THE STATE OF ART

In this section, the features of the current state-of-art system (highlighted inside the broken blue line in Fig. 1) and limitations (highlighted inside the broken red line in Fig. 1) are explained. Liu et al. proposed a novel conditional (CoNERF) for Facial Emotion Recognition in an unconstrained environment [10]. Their proposed method classified facial expressions (fear, happy, sad, anger, surprise, and disgust) at a multi-angle with conditional probabilistic learning technique (Gaussian Mixture Model GMM). their method provides an average classification accuracy of 98.9% and 89.7% on BU-3DEF and LFW datasets, respectively. Moreover, the average processing time for performing a pose-aligned Facial Expression Recognition is about 113 ms. The recognition accuracy for images from the wild is about 15% lower than the dataset images used for testing. The motivation behind selecting [10] work as the state of the art solution is that their proposed CNN model can learn deep high-level features that are extracted from salient facial patches by suppressing the influence of illumination, occlusion, and low image resolution from fewer numbers of training images whereas most of the other state-of-art solution considerable amount of image set for training. This model consists of two major stages, (a) deep salient feature representation and (b) post-aligned facial expression. See Fig. 1 and Table 1.

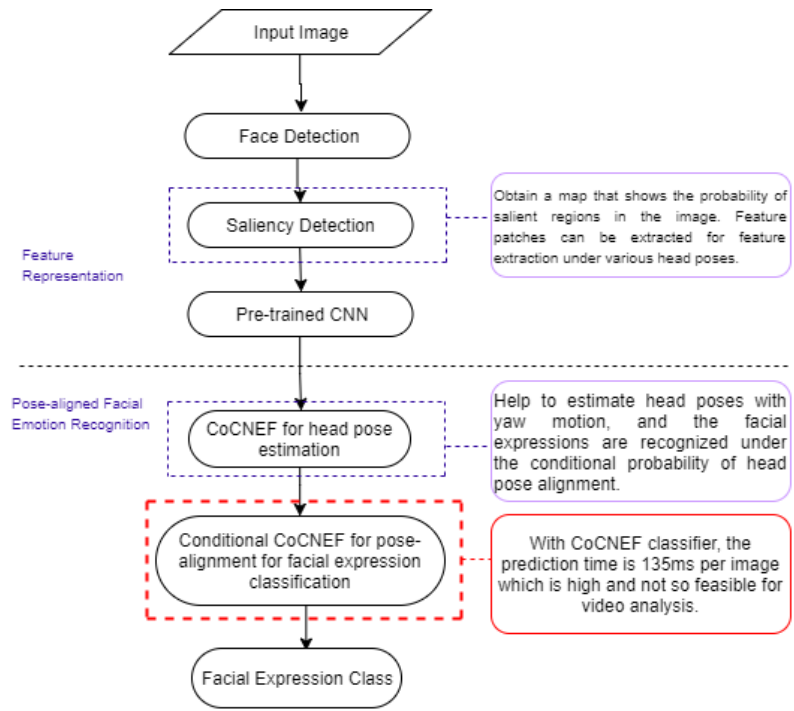

Fig. 1: Flow diagram of State of the Art System (Liu et al., [10]). The dotted blue box represents the good feature, and the red box represents the limitation.

*Deep Salient Feature Representation:* This stage starts from detecting the face and extracting the salient deep features from saliency-guided facial patches with pre-trained CNN model. The saliency detection algorithm has been adopted to sample salient face patch that distinguishes the salient region in the face depending on the image signature which contains information about the foreground of the image and also shows the probability of that region in the image. The image signature is computed by DCT in the algorithm, underlying the usefulness of the descriptor for detecting salient image regions. The salient region includes the eyes, nose, mouth, and cheeks. The similarity between facial patches has been measured to sample more distinct facial patches. The pre-trained VGG model is retrained with all the facial patches from LWF and YTF datasets to obtain deep high-level features. These features will be later matched with the image obtained from the camera (webcam) to detect the facial expression at real-time. However, in this model, the salient features are extracted directly from the input images without preprocessing stage. Raw images can contain unwanted features that may cause an error in salience detection algorithm results. The image has to be purified using PCA analysis before feeding to the pre-trained CNN model. With image resizing to 48x48 for dimension reduction (a pre-processing technique), the computational time is also reduced.

*Pose- aligned facial expression recognition stage:* In this stage, saliency-guided facial patches are used to extract the deep salient features that were detected in the previous stage under different head poses. Liu et al. [10] utilized Neurally Connected Split Function (NCSF) to improve the learning capability of decision tree nodes with the splitting child nodes by deep learning representation with CoNERF method. Each decision node is split into the leaf node unless each leaf contains the label information for one facial expression. Stochastic Gradient Descent (SGD) approach has been employed with the loss function minimize the risk of false feature representation in leaf nodes. The risk is minimized by adapting a random-set of feature samples from different leaf nodes. These nodes are grouped as a mini-batch frame of samples, that is used to train the VGG model. Each mini-batch of images contains seven facial expressions where the total facial expression has five sets, for a total of 35 images per batch. The SoftMax classifier, which is a generalization to multiple classes, is used in the output layer for the classification of facial expressions. The performance of the system can be improved with the replacement of this classifier with a fast Extensive Learning Machine (ELM) classifier that has fewer numbers of hidden layers. This classification layer has been implemented to bring down the processing time by Kaya et al. [20].

*Mathematical Approach:* The Stochastic gradient descent technique is employed to improve facial expression representation accuracy by minimizing the risk concerning Y in equation (1) [10] that reduces cross-entropy loss $L(Y, \pi; P)$ for the training sample P represented by loss function in equation (2) [10]. However, expressions can be described more accurately with the log-likelihood estimation technique that estimates the parameters of our proposed model.

Nitesh Banskota, Abeer Alsadoon, P.W.C. Prasad, Ahmed Dawoud, Tarik A. Rashid, Omar Hisham Alsadoon (2022). A novel enhanced convolution neural network with extreme learning machine: facial emotional recognition in psychology practices. Multimedia Tools and Applications. DOI : 10.1007/s11042-022-13567-8$$Y^{(t+1)} = Y^{(t)} - \frac{\eta}{|B|} \sum_{(P,\pi)\epsilon B} \frac{\partial L(Y,\pi;P)}{\partial Y} \tag{1}$$

and,

$$L(Y, \pi; P) = -\sum_n p(\pi|d_n, Y, P) \log(p(\pi|d_n, Y, P)) \tag{2}$$

Where,

Y = parameterization of network
η = learning rate
B = mini-batch of sample
L(Y, π; P) = log-loss term
P = training sample (salient feature sample)
π = facial expression label
t = position of Y
p(π|$d_n$, Y, P) = facial expression probability

The final facial expression probability is computed by a weighted average in equation(3) [10],

$$p(\pi| P ) = \frac{1}{T} \sum_i \sum_{t=1}^{k_i} p(\pi|l_{t,\Omega_i}(p)) \tag{3}$$

Where,

t = total number of samples
π = facial expression label
i = number of iterations
$l_{t,\Omega_i}$ = corresponding leaf for feature representation from the decision tree.

NCFS-$f_n$ has been introduced to improve the learning capability of a splitting node by deep learning representation. Each splitting node output is determined with respect to NCSF function as equation (4) [10],

$$d_n(P, Y|\theta) = \sigma(f_n(P, Y|\theta)) \tag{4}$$

Where,

P = training sample (salient feature sample)
Y = parameterization of network
θ = Head pose

Here,

$$\sigma(x) = (1 + e^{-x})^{-1} \tag{5}$$

is a sigmoid function.

x = input variable that represents NCFS function

To address the Multiview face-positions, convolution CoNERF is trained to estimate head poses in nine yaw axis -90o to +90o at 15o difference. The extracted features are nine probabilistic models for head posed are combined by adopting a Gaussian Mixture model, which is shown in equation(4) [10],

$$p(\theta|l) = N(\theta; \bar{\theta}, \sum_l^\theta) \tag{6}$$

Where,

$\theta$ = mean of head pose probabilities.

$\sum_l^\theta$ = Covariance of head pose probabilities

Table 1: Algorithm for facial emotion recognition in state-of-art

| **Algorithm**: Conditional convolution neural network enhanced random forest for facial expression recognition |
|---|
| **Input**: CK+, JAFFE, multi-view BU-3DEF and LFW datasets |
| **Output**: Improved Facial Emotion Recognition model |
| BEGIN<br>    1. Deep salient feature extraction from facial patches<br>    2. Detect of head pose 0<br>    3. Splitting the nodes by NCSF<br>    4. Create leaves nodes<br>    5. Obtain the probability of the leave nodes<br>    6. Classify image feature with CoNERF model<br>END |

## 4. PROPOSED SYSTEM

After analyzing several existing methods for Facial Emotion Recognition with Deep Learning Technique, we evaluate the merit and demerits of each method. Accuracy, processing time, image illumination, feature extraction and head-pose alignment [4] [10-16] were the main issues to be considered. From the collected list of reviews, we selected the method proposed by Liu et al. [10] as the best solution since it extracts deep salient features from saliency-guided facial patches and uses a probabilistic learning method based on (GMM) for facial expression recognition to achieve an accurate outcome. This method represents and manipulates uncertainty from input data and prediction's Log Loss function. The sigmoid activation function adapted by Liu et al. [10] is best suited for predicting facial emotions for six different emotion classes due to the obtained accuracy and the processing time comparing to other systems. Moreover, some modifications in the classification stage can be implemented on this state of art solution with the help of selected second-best solution proposed by Kaya et al. [20] using (ELM) classifier instead of SoftMax classifier. See fig.2

Kaya et al. [20] implemented Linear and Radial Basis Function (RBF) kernels of ELM classifier to enhance the learning capabilities of the system. ELM was initially proposed for generalized single-hidden-layer feedforward neural networks and overcome the local minima, learning rate, stopping criteria and learning epochs that exist in gradient-based methods such as back-propagation (BP). ELMs are widely used due to some significant advantages such as learning speed, ease of implementation and minimal human intervention. The potential for large-scale learning and artificial intelligence is preserved. The main steps of ELM include the random projection of a hidden layer with random input weight's algorithm. Since the number of layers is only two in ELM classifier, the processing time is highly reduced because the required learning times for two layers are less than multiple layers. This method has a negligible impact on the accuracy of the system because CNN is used for feature extraction, which is the main advantage of CNN[19]. This approach has the capability to bring down the processing time of 113ms of the state of art method to around 65ms, which is suitable for expression detection in video analysis.

Additionally, an entire Optical Flow Estimator and pre-processing of image blocks are proposed in the feature representation stage to prepare clear images without noise and fewer pixels for the feature extraction stage. It controls the video frame flow rate to obtain an image for further processing and utilizes a Principal Component Analysis (PCA) technique in fig.3 for false-positive face detection by measuring the mean reconstruction error per image, after projecting the images to the PCA space and back. The frames with a high reconstruction error are discarded, as these are probably poorly detected or poorly aligned images. These techniques have been used in the second-best solution by Kaya et al. [20].

The proposed system consists of three major stages (Fig. 2), (a). Pre-processing, (b). Feature Extraction, and (c). Classification. The stages in state of art solutions are modified as these three stages with some additional steps as shown in fig.3.

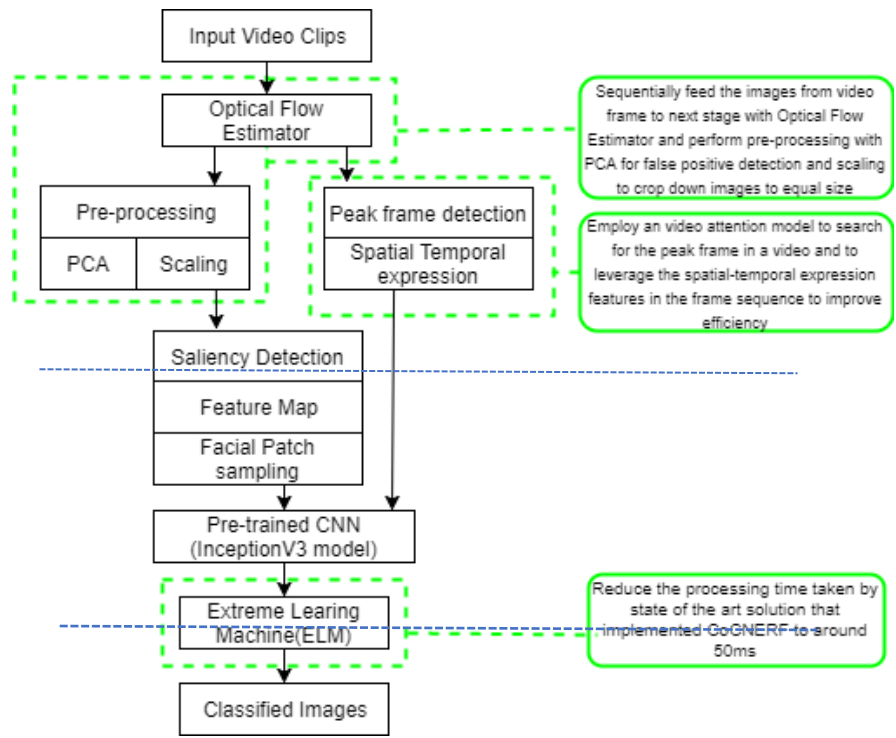

Fig. 2: Block Diagram of Proposed Solution. Green represents the additional proposed steps to state of art solution

*Preprocessing:* In this stage, the images are obtained from video clips. These images are sequenced, and their flow rate is controlled with the optical flow estimator so that the peak frame from the video clip is made ready for pre-processing. The image thus obtained is converted into a grayscale image. PCA analysis which measures the mean reconstruction error per image, after projecting the images to the PCA space and back, is done to detect the false-positive faces so that those false images with a high reconstruction error, as these are probably poorly detected or poorly aligned can be discarded [20]. The image is then scaled to 48x48 to get a unified features vector size in the feature extraction stage. As the state of art solution reviewed above, the images were not pre-processed, and the features are extracted directly from the obtained images. In this proposed paper, the images are purified with PCA so that the transfer learning process in the proposed CNNEELM model could be established efficiently.

*Feature Extraction:* The salient region in the face (eyes, nose, cheeks and mouth) is detected and nine salient facial patches are created. These patches highlight sparse salient regions based on image signature and contain features that are to be extracted in the next stage. These patches are fed to pre-trained CNN, which represents the features in some numerical value. In the proposed system, Inception v3 is a widely-used image recognition model that is made up of symmetric and asymmetric building blocks, including convolutions, average pooling, max pooling, concerts, dropouts, and fully connected layers. InceptionV3 model has been used instead of VGG face pre-trained model because of its easy availability and less complexity in convolution layer alignments. Simultaneously, with the peak frame detected by the optical flow estimator, the video attention model is employed to leverage the spatial-temporal expression features in the frame sequence to improve efficiency [10]. From this frame, the feature is extracted with Fisher Vector (FV) encoding technique proposed by kaya et al. [20], which is not a deep learning technique. FV provides a supra-frame encoding of the local descriptors, quantifying the gradient of the parameters of the background model with respect to the data.

The state of art solution did not employ the video attention model; thus, the fusion of deep learned network and machine learning was not important. The proposed system fuses these two methods to improve the accuracy in feature representation.

*Classification:* For classification, the Softmax classifier of CNN, which use the 6-dimensional expression probabilities for prediction, is replaced with the ELM classifier layers for better predictive analysis. The advantage of ELM classifier is that it can classify the features very quickly into different classes. Proposed by Huang et al. [21], ELM classifier is a feedforward neural network with single hidden layer feedforward networks (SLFNs) nods. Sic it used only SLFNs nods the ELM classifier is very popular, which boosts up the training speed and reduces overfitting and prediction parameters to a new plane to determine linear regression. . In the proposed system, ELM classifier is used instead of SoftMax classifier used by Liu et al. [10] in order to get the best possible results.

**Proposed Equation:**

The stochastic gradient descent SGD technique is employed with the loss function in equation (1) with the log-loss term as equation (2) [10]. SGD can converge faster than batch training of extracted features because it performs updates more frequently. Mini-batch samples of features for training are major characteristics of SGD. This can be faster than training on single data points because it can take advantage of vectored operations to process the entire mini-batch at once. The performance of SGD can be improved with the replacement of Log loss function [10].

The log value of the probability function gives the negative result which is neutralized by the negative sign to make the calculation easier. However, equation (2) does not smoothen the stochastic gradient descent curve from equation (5), which may result in some predictions being unconsidered within the curve.

**Log Loss** calculates the uncertainty of the prediction depending on how much it varies from the original input. However, in our proposed system, the log-loss term is entirely replaced with the maximum likelihood function in order to avoid the problem of overfitting. So, the function could give the maximum likeliness of prediction to fall under the curve. The modifying log-likelihood estimator($L_I(Y, \pi; P)$) function is shown in equation (7) [22].

$$L_I(Y, \pi; P) = \frac{1}{n}\sum_{i=1}^{n} Log(p^{\Sigma Y_i}(1-p)^{n-\Sigma Y_i}) \quad (7)$$

Our Modified Stochastic gradient descent for maximum likelihood estimator function for loss is presented in equation (8)

$$Y_M^{(t+1)} = -Y_M^{(t)} * \frac{\eta}{2|B|} \sum_{(P,\ \pi)\epsilon B} \frac{\partial L_I(Y,\pi;P)}{\partial Y} \quad (8)$$

Where,
$L_I$ = Likelihood estimator
n = total number of samples
p = facial expression probability
x = input variable

In the above equation (8), it can be observed that the subtraction of variables inside summation in equation (5) is replaced with the multiplication sign so that value of the new function can be plotted in the logarithmic curve, which makes the proposed prediction model more accurate. The negative sign is introduced at the beginning of the right-hand side of equation (8) to neutralize the negative parameterization of the log probability function. Additionally, the batch size is doubled, resulting in minimization of processing time during the training session.

The activation function named sigmoid function in equation (4) in the state of art solution is also modified by us as equation (9).

$$\sigma_M(x) = (1 + e^{-\frac{x}{2}})^{-1} \quad (9)$$

Hence, the modified NCSF function will be presented as equation (10).

$$d_{n_M}(P, Y|\theta) = \sigma(f_n \frac{(P,Y|\theta)}{2}) \quad (10)$$

The exponential curve by sigmoid function is "S" shaped curve whose shape can be made flatter by multiplying the input parameter in the right-hand side of equation(9) by half to obtain less varied output for rapid change in input. Towards the end of the sigmoid curve in equation (4), there exists a vanishing point at which the change in the input does not cause the output to change. In order to push back the vanishing point further away, the proposed curve is made slightly flatter so that feature representation is done at a slower rate to improve accuracy in feature representation and a classification stage. This modification causes a negligible increase in processing time. We propose to take this modified function as our reference function to make a prediction for feature representation for six different classes.

We propose the modification of two equations (1), (4) from the state of art solution proposed by Liu et al. [10]. First, the risk in feature extraction is high due to cross-entropy loss in Stochastic Gradient Decent approach. With the help of the proposed equation (8), the likelihood of feature prediction is improved at each iteration, which minimizes the risk factor. Along with this, the sample size is doubled so that the system can be trained at a quicker rate. Second, the shape of the activation function is altered and has been made flatter with equation (9). The enhanced activation sigmoid curve gives less varied output for rapid change in input. This results in a slower rate of learning at the training stage so that the system can achieve more accurate results during facial expression detection.

Extreme Learning Machine(ELM) can easily classify the one-dimensional vectors which are obtained from the feature map of Convolution Neural Network fused. The facial expression features extracted from CNN is used to create a facial map used for gathering all the obtained features. This map is utilized by ELM to compute the hidden layer weight to get the best possible one with random input from the fully connected layer of CNN. Since ELM has only one hidden layer in its structure, the classification process is swift. Additionally, ELM maintains the feature generalizing ability at the reduced computational cost of classification under the proper selection of activation function (in our case, sigmoid function). Thus, this classifier is very suitable for emotion detection in video clips as it is twice as fast as CNN classifiers [23]. The literature review in this paper shows that none of the authors has considered fusing CNN with ELM classifier to classify facial emotion. The proposed solution reduces the processing time without affecting the accuracy of feature extraction by CNN. However, the proposed CNNEELM model for facial emotion recognition is illustrated in Table 2.

Table 2: Algorithm of proposed CNNEELM model for facial emotion recognition

| Algorithm: Proposed CNNEELM model for facial emotion recognition |
|---|
| Input: Image containing facial image at real-time, CK+ dataset, JAFFE dataset and FER2013 dataset |
| Output: Facial Emotion Recognition |
| START<br>  1. Feed the system with video frames<br>  2. Detect peak frame<br>      i. If peak frame, advance to step 3<br>      ii. Else, go to step 2<br>  3. Extract Features<br>      i. Extract salient features with a Saliency detection algorithm<br>      ii. Obtain spatial-temporal expression<br>  4. Check whether the features are from the same peak frame,<br>      i. If no, discard features and go to step 2<br>  5. Combine features with CNN<br>  6. ELM classification<br>  7. Obtain Facial Emotion recognition<br>END |

## 5. IMPLEMENTATION OF THE MODEL

Our evaluation was implemented in Python 3.7.1 programming language with OpenCV and Tensorflow libraries. Three datasets CK+ [24] JAFFE [25] and FER2013 was used to train and validate the model. 30,709 images in total were used to train the system. CK+ dataset contains eight emotional expression images from 123 subjects. There is a total of 593 image sequences, and most of the images are grey. The JAFFE dataset contains 6 peak emotions with 1 neutral expression class. Total of 10 Japanese models' facial expression is captured consisting of 213 static images. Fig.7 and Fig.8 shows the number of images in each category for CK+ and JAFFE dataset. Five images from for six different classes (Happy, sad, surprise, disgust, fear and neutral) from CK+ has been randomly taken as a sample to validate the proposed model. The images of men and women of different ages are considered. The resolution of images is different for each dataset. For CK+ dataset the image resolution is 720x480 or 460x480; for JAFEE, it is 256x256 and for FER2013, it is 48x48 due to the video quality. The images are categorized into six different folders - happy, sad, surprise, disgust, neutral and fear with Python.

In CK+ dataset, the image sequence number was matched with their respective labels in label.txt file and were categorized with python code. The JAFFE dataset contained images, and their filename had emotion labels on them. The filename was split to get the category and images were classified. In the case of FER2013 dataset, the raw images were represented in the CSV file. There is total of 35,887 grayscaled images. The file contained three columns – emotion (0-5), image data and set(validating set and training set). The CSV file was converted to NumPy(.npy) file in Python which could be used directly to train the model. The images from all the datasets were pre-processed with PCA analysis technique where all faces are scaled at 48x48 images and converted to grayscale.

For video analysis, an image sequence from CK+ dataset, at 23fps, showing different facial emotions were fed into a system to determine the processing time of the system. The video length is only 10 seconds. In order to calculate the processing time, the time at which frame in input to the system is subtracted from output is computed. Some of the frames that are irrelevant to the experiment has been ignored.

In pre-processing stage of the proposed model, the face is detected with the Haar cascade classifier. The summary of four different face detection techniques from Haar cascade module is used to avoid false face detection, whereas the state of art model does not consider such an issue. The facial region is converted into facial patches, showing only the eyes, nose, cheek and mouth with the saliency detection algorithm. The image is then sampled into 9 different patches, each patch containing facial features as shown in fig.3. For ease, only one head pose, that is, the frontal view, is considered for the experiment. These facial patches are then fed to pre-trained model InceptionV3 for facial expression feature extraction and representation. The output layer with SoftMax function is modified and replaced with ELM layer to decrease the processing time during classification.

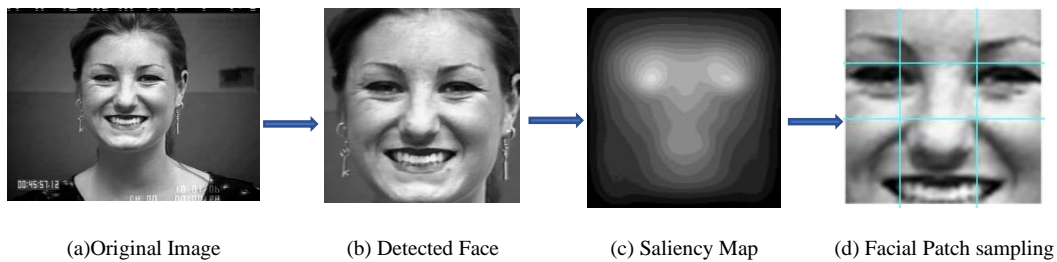

(a) Original Image    (b) Detected Face    (c) Saliency Map    (d) Facial Patch sampling

Fig. 3: The salient patch sampling in a face

## 6. RESULT

In order to understand the influence of accuracy and processing time, we conducted an experiment with deeply extracted features from CNN and ELM for classification and compared the results with the state of art solution. From Table 3, it can be observed that in most of the cases, the accuracy is better than the results obtained from state of art. Additionally, the processing time in the proposed solution is almost half of the state of art processing times. Here, we have been able to bring the processing time from around 113 ms to 60ms on average. This results in better system response to video data.

All the samples for experiments are taken from all three datasets, with each category containing five images from CK+ database. With the help of graphs and tables, the proposed system is compared with the state of art. The results of the experiments performed on few samples for facial expression classification are listed in Table 3. For feature extraction, during the training session, the bottleneck files are created in text format with an index number so that a dictionary of training images could be created. In Python, a graph is created that returns the graph object. The graph holds the trained network and other various tensors. The pre-trained InceptionV3 network is trained with the bottleneck files. The network is updated with new files that plot a new prediction graph on output graph file in .pb format. This graph is used later during image recognition to classify the facial image. In testing samples in Table 3, there are cases where the change in accuracy is significantly high in some cases and similar in others. Considering sample 3.4 from Table 3 for sad face class, the original image is a little blurred due to the image noise which affected the accuracy in recognition in the state of art method. Our proposed model pre-processed the image with PCA analysis, which discards the images with high reconstruction error, improving the lighting intensities of the pixel, hence improving the accuracy. See Figures (4-13)

Fig.6, 7, and 8 show the average accuracy of facial expression classification in all image categories (happy, sad, disgusted, fear, neutral, surprise and angry) in CK+, FER2013 and JAFFE datasets, respectively. The graph shows that the accuracies for each emotion category are in the range of 85% to the high end of 90% for the various selected emotions. In all three datasets, the accuracy for 'disgust' and 'fear' is less as compared to other categories because of the similarity in extracted facial features. 10% of the images selected randomly from each dataset were separated before training the model. These images are used for testing the accuracy and processing time of the proposed model. The average of accuracy and processing time for each sample is computed to obtain the overall accuracy and processing time in all the datasets. Fig.9 shows that the overall accuracy has been improved in the proposed solution. The processing time has been reduced almost by half after adapting the ELM classification layer which is shown in fig.10.

During the training stage, the images are scaled to 48x48 and only salient features are detected. The benefit of pre-processing can be observed during the training session. While training, the system learns only the specific area in the image, i.e., eyes, nose, cheeks and mouth. This results in faster training since the model needs to process fewer image pixel values in the classification stage. During feature extraction and representation, the dataset is categorized into three sets, training, validation set and testing set. The pre-trained CNN model is retrained with a training set with the unsupervised learning method. To control the overfitting and maintain regularization, the validation set is used to check whether the model is learning positively or not. Fig.11 shows the smoothened plot of the accuracy of training and validation of image classification. The fig. shows that as the system is trained with facial images, the accuracy is low for both the training set and validation set in the beginning but at the end, the validation accuracy is above 95% and is consistent. As the number of epochs increases the cross-entropy for training is decreasing and has reached less than 0.1, whereas, for validation, the cross-entropy is around 0.3.

All these values are calculated in the TensorFlow library. While developing the system, the tensors are attached with summary modules, which log all the necessary parameters required to generate a graph during and after the training session. The summary log includes curves for training and validation accuracy during training sessions (fig.11), cross-entropy loss (fig.12) and standard deviation, mean for bias values and weights (Fig.13). These curves are plotted in TensorBoard, which comes pre-installed in TensorFlow library. The smoothness of the curve is made high so that the change in the graph is distinctively visible.

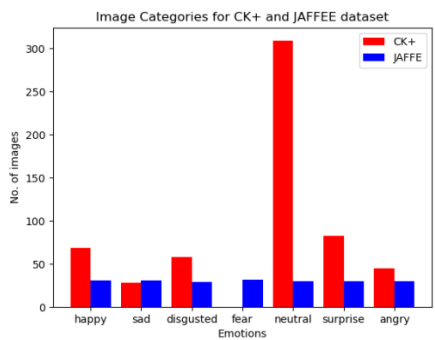

Fig. 4: total number of images for CK+ and JAFFE dataset for the training module

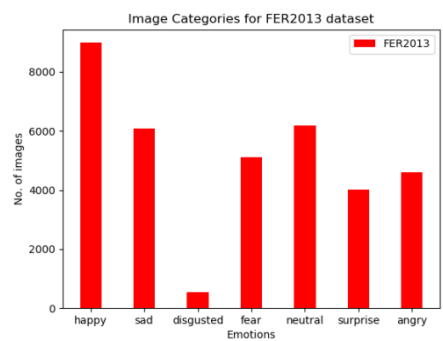

Fig. 5: Total number of images for FER3013 dataset for the training module

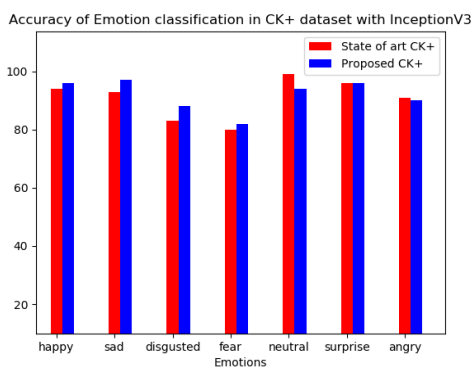

Fig. 6 Accuracy of Emotion Classification in CK+ dataset with InceptionV3

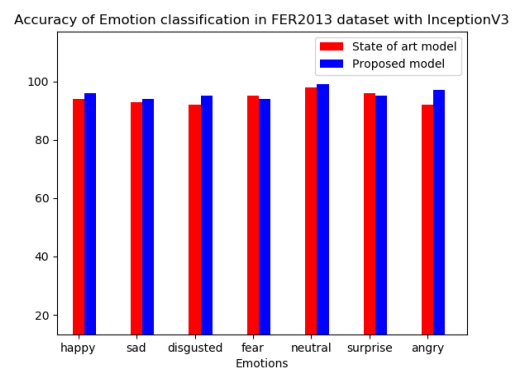

Fig. 7 Accuracy of Emotion Classification in FER2013 dataset with InceptionV3

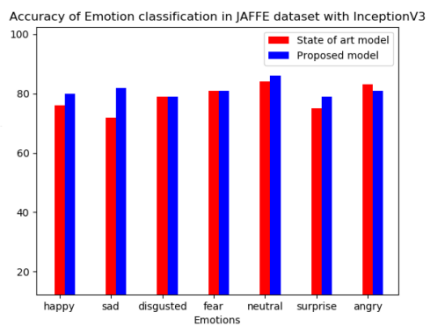

Fig. 8 Accuracy of Emotion Classification JAFFE dataset with InceptionV3

Fig. 6-8: Comparison of classification accuracy between state of the art and proposed solution in (fig 6) CK+ dataset, (fig 7) FER2013 dataset and (Fig 8) JAFFE dataset for each emotion – happy, sad, disgusted, fear, neutral, surprise and angry.

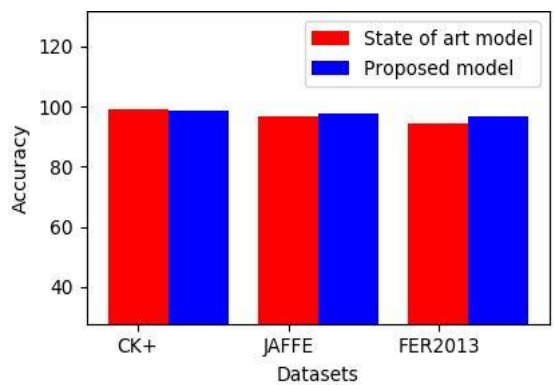 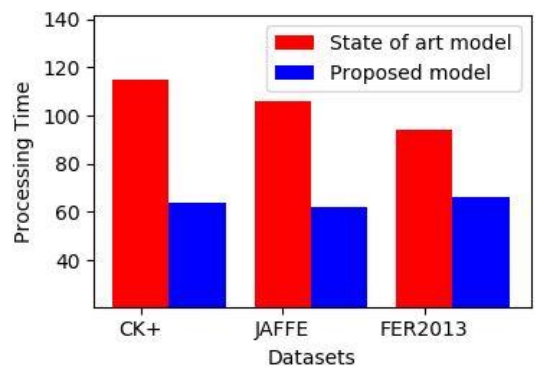

Fig. 9: Overall average accuracy of the state of art method and proposed system

Fig. 10:Overall processing time(in ms) for state of art method and proposed system

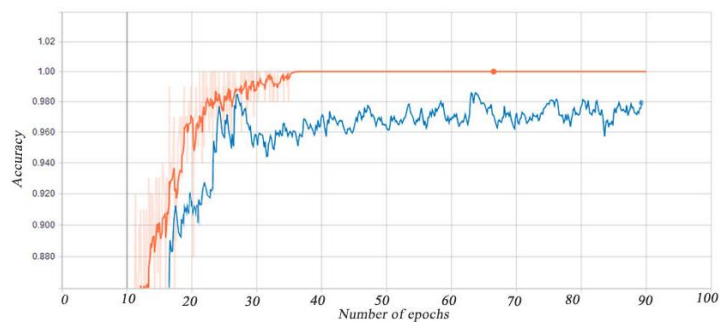

Fig. 11: Graph plot for 100 epochs showing Training accuracy (red) Vs. Validation accuracy (blue)

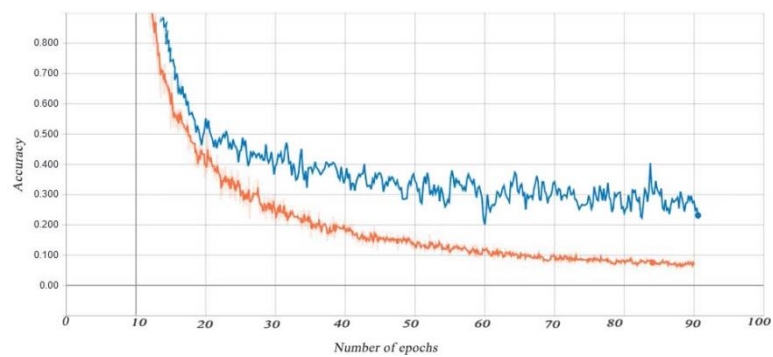

Fig. 12: Cross Entropy loss during training session for 100 epochs for Training set (red) and Validation set (blue)

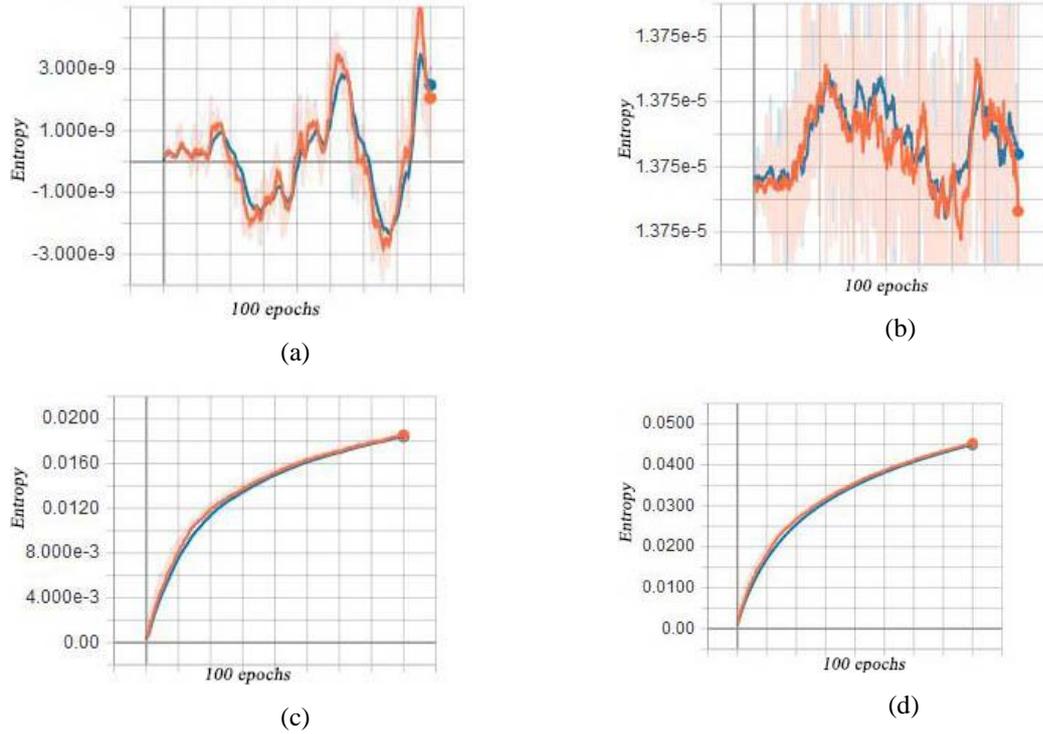

Fig. 13: Graph plot for Bias and weight summaries during a training session for 100 epochs for training (red) and validation(blue) set (a) Bias summaries for the mean (b) Mean Weights summaries (c) Bias summaries for standard deviation (d) Weights summaries for standard deviation

Table 3: Accuracy and Processing Time result for CK+ dataset

| Emotion No. | Original image | State of the art | | | Proposed solution | | |
|---|---|---|---|---|---|---|---|
| | | Processed image | Accuracy in terms of cross-entropy loss (%) | Processing Time(ms) | Processed Image | Accuracy in terms of cross-entropy loss (%) | Processing Time(ms |
| 1. Sample Images for Neutral Face | | | | | | | |
| 1.1 | | | 94.55 | 115 | | 96.55 | 60 |
| 1.2 | | | 98.84 | 113 | | 98.12 | 65 |
| 1.3 | | | 85.85 | 123 | | 89.33 | 64 |
| 1.4 | | | 99.54 | 121 | | 97.66 | 78 |
| 1.5 | | | 92.75 | 102 | | 93.56 | 59 |
| 2. Sample images for Happy Face | | | | | | | |
| 2.1 | | | 99.87 | 112 | | 98.03 | 63 |
| 2.2 | 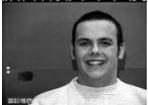 | 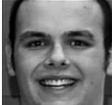 | 93.41 | 105 | 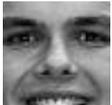 | 97.54 | 65 |
| 2.3 | 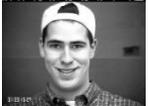 | 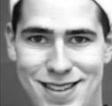 | 92.76 | 113 | 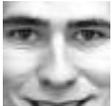 | 93.65 | 67 |
| 2.4 | 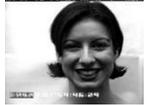 | 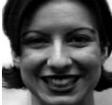 | 96.47 | 115 | 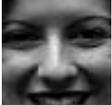 | 96.78 | 65 |
| 2.5 | 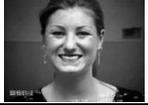 | 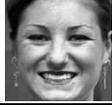 | 98.79 | 132 | 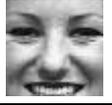 | 98.56 | 63 |
| | 3. Sample Images for Sad Face | | | | | | |
| 3.1 | 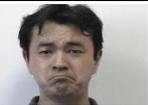 | 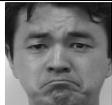 | 94.55 | 120 | 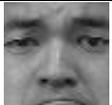 | 95.95 | 61 |
| 3.2 | 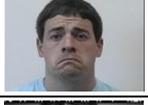 | 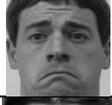 | 95.07 | 98 | 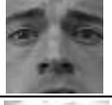 | 92.44 | 71 |
| 3.3 | 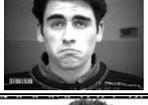 | 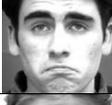 | 97.80 | 99 | 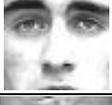 | 92.31 | 70 |
| 3.4 | 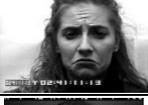 | 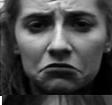 | 83.54 | 98 | 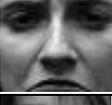 | 99.50 | 68 |
| 3.5 | 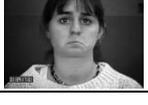 | 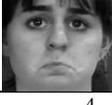 | 88.46 | 111 | 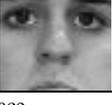 | 93.33 | 59 |
| | 4. Sample images for Angry Face | | | | | | |
| 4.1 | 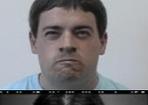 | 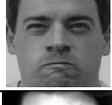 | 93.86 | 105 | 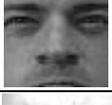 | 94.65 | 58 |
| 4.2 | 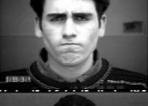 | 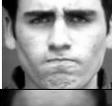 | 87.36 | 106 | 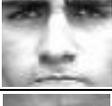 | 91.65 | 59 |
| 4.3 | 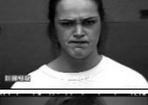 | 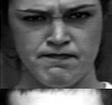 | 97.65 | 112 | 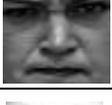 | 98.76 | 64 |
| 4.4 | 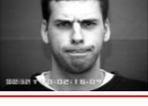 | 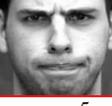 | 89.47 | 97 | 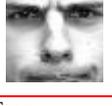 | 86.88 | 67 |
| | 5. Sample Images for Neutral Face | | | | | | |
| 5.1 | 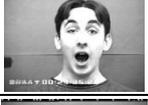 | 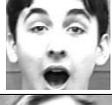 | 98.96 | 107 | 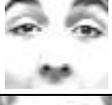 | 95.76 | 66 |
| 5.2 | 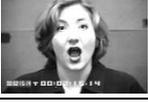 | 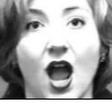 | 97.34 | 103 | 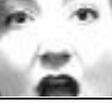 | 98.68 | 61 |

| 5.3 | 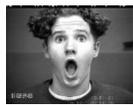 | 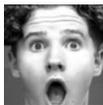 | 93.75 | 108 | 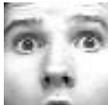 | 92.76 | 60 |
|---|---|---|---|---|---|---|---|
| 5.4 | 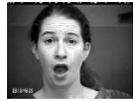 | 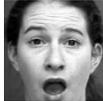 | 96.37 | 105 | 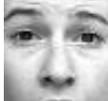 | 97.33 | 65 |
| 5.5 | 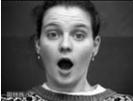 | 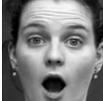 | 92.21 | 104 | 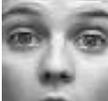 | 95.78 | 67 |
| 6. | | | Sample images for Disgust Face | | | | |
| 6.1 | 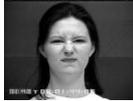 | 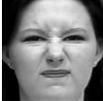 | 85.15 | 108 | 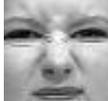 | 91.43 | 63 |
| 6.2 | 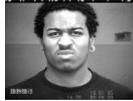 | 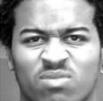 | 82.36 | 113 | 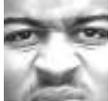 | 86.30 | 70 |
| 6.3 | 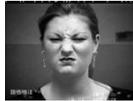 | 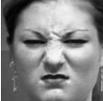 | 78.27 | 112 | 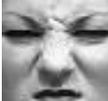 | 88.64 | 72 |
| 6.4 | 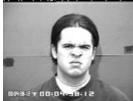 | 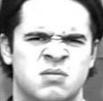 | 80.54 | 119 | 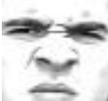 | 90.43 | 77 |
| 6.5 | 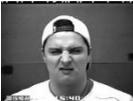 | 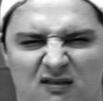 | 88.67 | 116 | 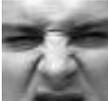 | 91.76 | 71 |

## 7. DISCUSSION

The comparison between state of the art and the proposed system shows the overall improvement inaccuracy in terms of image registration using the proposed CNNEELM model during the training session of the detected features. Likewise, the image-processing time of classification has been substantially decreased (as shown in table 3) during classification within fast ELM classifier hence increasing the system performance. The proposed system with enhanced equation (8) improves the overall accuracy during image registration by roughly 2%. The difference between training accuracy and validation accuracy is minimal as shown in Fig.11, which means that no overfitting has occurred in the classification stage. During classification, the image frames are classified at around 20 fps for better recognition results, which show that the system can be adopted emotion recognition in video clips as well. These parameters are simulated in TensorFlow library by implementing both state of art and proposed systems. We compute the system recognition accuracy as in Fig.9by calculating the mean of log-likelihood estimation in the activation layer of the convolution neural network.

The main feature of the proposed system is the influence on the accuracy during image recognition as shown in Table 3. The minimization of the risk factor of overfitting due to the adoption of the maximum-likelihood estimator in SGD provides an increase in accuracy by 2% during feature recognition than the state of art system. Zhao et al. [22] implemented a similar estimator to a study of bone marrow transplantation for patients with acute leukemia. ELM classifier has proven to be more advantageous in terms of processing time than traditional SoftMax classifier since it reduced the processing time by half than state of the art. During the pre-processing stage, both state of art and the proposed method use saliency detection technique in a similar fashion. The patch size of extracted features in the proposed system is 12x12 resolution, which is slightly bigger than state-of-the-art so that no features are ignored during facial patch sampling.

## 8. CONCLUSION

Human facial expression provides an essential clue to understanding and analyzing human emotion and their behaviours. The proposed method utilizes the full potential of Convolution Neural Network enhanced with modified stochastic gradient descent for maximum likelihood estimator function. The pre-processing of training and testing dataset has proven to provide extra efficiency in feature representation. The additional ELM classifier successfully classified the extracted facial features of a human face with high precision at minimal processing time as illustrated in Table 3. The experiment result shows that the proposed method is effective and robust than the state of art solution in the aspect of accuracy and the processing time. The average accuracy of image recognition in CK+, JAFFE and FER2013 datasets is 98.40%, 97.60% and 96.53%, respectively. The time of performance on a facial image is about 65ms per image. For video analysis, with the introduction of optical flow estimation technique, the facial emotion can be detected in real-time for video up to 25fps.

In conclusion, ELM reduces the computational cost without degrading the generalization capability under the proper selection of activation function, while the state of art system requires much computational power to classify the facial features. In addition to, Partial Least Square regression in our proposed solution reduces the number of predictors so that the variance can be minimized during classification. The state of the art system does not address the variance in classification due to the unidentified number of predictors.

In the future, the proposed model can be modified to detect the micro-expression by considering each frame from the video clip for extracting the desired features rather than the peak frame. Additionally, motion blur images can be pre-processed for low-rank matrix reconstruction for better recognition results.

## DECLERATION

No Funding for this work and no Conflicts of interests as well

## Appendix

List of Abbreviation

| | |
|---|---|
| **CK** | Cohn-Kanade |
| **CNN** | Convolution Neural Network |
| **CoNERF** | Conditional Convolution Neural Network Enhanced Random Forest |
| **ELM** | Extensive Learning Machine |
| **FER** | Facial Emotion Recognition |
| **JAFFE** | Japanese Female Facial Expression |
| **NCSF** | Neurally Connected Split Function |
| **PCA** | Principal Component Analysis |
| **PLS** | Partial Least Square |
| **RF** | Random Forest |
| **SGD** | Stochastic Gradient Descent |
| **SIFT** | Scale-Invariant Feature Detection |
| **VGG** | Visual Geometry Group |
| **GMM** | Gaussian Mixture Model |

## DECLERATION

No Funding for this work

No Conflicts of interests for this work

Nitesh Banskota, Abeer Alsadoon, P.W.C. Prasad, Ahmed Dawoud, Tarik A. Rashid, Omar Hisham Alsadoon (2022). A novel enhanced convolution neural network with extreme learning machine: facial emotional recognition in psychology practices. Multimedia Tools and Applications. DOI : 10.1007/s11042-022-13567-8

No Real Data has used